\documentclass{article}

\usepackage{arxiv}

\usepackage[utf8]{inputenc} 
\usepackage[T1]{fontenc}    
\usepackage{hyperref}       
\usepackage{url}            
\usepackage{booktabs}       
\usepackage{amsfonts}       
\usepackage{nicefrac}       
\usepackage{microtype}      
\usepackage{lipsum}
\usepackage{graphicx}
\usepackage{amsmath,amssymb}
\usepackage{lmodern}
\usepackage{iftex}
\graphicspath{ {./images/} }

\usepackage{color}
\usepackage{fancyvrb}
\usepackage{etoolbox}
\usepackage{calc} 
\usepackage{booktabs}
\usepackage{longtable}
\usepackage{array}
\usepackage{multirow}
\usepackage{wrapfig}
\usepackage{float}
\usepackage{colortbl}
\usepackage{pdflscape}
\usepackage{tabu}
\usepackage{threeparttable}
\usepackage{threeparttablex}
\usepackage[normalem]{ulem}
\usepackage{makecell}
\usepackage{xcolor}

\DefineVerbatimEnvironment{Highlighting}{Verbatim}{commandchars=\\\{\}}
\usepackage{framed}
\definecolor{shadecolor}{RGB}{241,243,245}
\newenvironment{Shaded}{\begin{snugshade}}{\end{snugshade}}

\newcommand{\AttributeTok}[1]{\textcolor[rgb]{0.40,0.45,0.13}{#1}}

\newcommand{\ConstantTok}[1]{\textcolor[rgb]{0.56,0.35,0.01}{#1}}
\newcommand{\ControlFlowTok}[1]{\textcolor[rgb]{0.00,0.23,0.31}{#1}}

\newcommand{\DecValTok}[1]{\textcolor[rgb]{0.68,0.00,0.00}{#1}}

\newcommand{\FloatTok}[1]{\textcolor[rgb]{0.68,0.00,0.00}{#1}}
\newcommand{\FunctionTok}[1]{\textcolor[rgb]{0.28,0.35,0.67}{#1}}

\newcommand{\NormalTok}[1]{\textcolor[rgb]{0.00,0.23,0.31}{#1}}

\newcommand{\OtherTok}[1]{\textcolor[rgb]{0.00,0.23,0.31}{#1}}

\newcommand{\SpecialCharTok}[1]{\textcolor[rgb]{0.37,0.37,0.37}{#1}}

\newcommand{\StringTok}[1]{\textcolor[rgb]{0.13,0.47,0.30}{#1}}

\providecommand{\tightlist}{%
  \setlength{\itemsep}{0pt}\setlength{\parskip}{0pt}}\usepackage{longtable,booktabs,array}

\title{EZtune: A Package for Automated Hyperparameter Tuning in R}

\author{
 Jill Lundell \\
  Department of Data Science\\
  Dana-Farber Cancer Institute\\
  Department of Biostatistics\\
  Harvard T.H. Chan School of Public Health\\
  Boston, MA 02215 \\
  \texttt{jlundell@ds.dfci.harvard.edu} \\
}

\begin{document}
\maketitle
\begin{abstract}
Statistical learning models have been growing in popularity in recent years. Many of these models have hyperparameters that must be tuned for models to perform well. Tuning these parameters is not trivial. EZtune is an R package with a simple user interface that can tune support vector machines, adaboost, gradient boosting machines, and elastic net. We first provide a brief summary of the the models that EZtune can tune, including a discussion of each of their hyperparameters. We then compare the ease of using EZtune, caret, and tidymodels. This is followed with a comparison of the accuracy and computation times for models tuned with EZtune and tidymodels. We conclude with a demonstration of how how EZtune can be used to help select a final model with optimal predictive power. Our comparison shows that EZtune can tune support vector machines and gradient boosting machines with EZtune also provides a user interface that is easy to use for a novice to statistical learning models or R.
\end{abstract}


\section{Introduction}
Statistical learning models provide powerful alternatives to more
traditional statistical models, such as regression. However, many of
these models have hyperparameters that must be tuned in order to achieve
optimal prediction accuracy. Many methods have been proposed for tuning
hyperparameters for statistical learning models, but few of these
methods are supported with research. The popular R packages \cite{R} \texttt{tidymodels} \cite{tidymodels} and \texttt{caret}
\cite{caret}, automatically tune hyperparameters, but they can be
prohibitively difficult to implement for a less experienced R user or
someone new to machine learning. We introduce a package called
\texttt{EZtune} that automatically tunes hyperparameters for support
vector machines (SVMs) \cite{svm}, gradient boosting
machines (GBMs) \cite{gbmpaper}, adaboost \cite{adaboost}, and elastic net \cite{en}. \texttt{EZtune} has a
simple user interface that is accessible to a novice R user, uses a
method to tune hyperparameters that is well documented, and its ability
to consistently tune an accurate model is backed by research \cite{dissertation}. First, we provide a short introduction to SVMs, boosted trees,
and elastic net with a focus on their respective hyperparameters. This
is followed by an overview of \texttt{EZtune}, \texttt{tidymodels}, and
\texttt{caret}. Next, we compare the performance of \texttt{EZtune} with
\texttt{tidymodels} and \texttt{glmnet} \cite{glmnet} for hyperparameter tuning. The No Free Lunch theorem
indicates that no one model type outperforms all other models in every
situation \cite{schumacher2001no}. Thus, we conclude with a
demonstration of how \texttt{EZtune} can be used to tune different
support vector machines, gradient boosting machines, and elastic net
models to select the model with the best performance.

\hypertarget{overtune}{%
\section{Overview of tuning parameters}\label{overtune}}

The following section briefly summarizes SVMs, boosted trees, and
elastic net and identifies the hyperparameters for each model. The focus
of each summary is the identification of hyperparameters that require
tuning for each model type.

\hypertarget{svmintro}{%
\subsection{Support Vector Machines}\label{svmintro}}

SVMs use separating hyperplanes to create decision boundaries for
classification and regression models \cite{svm}. The
separating hyperplane is called a soft margin because it allows some
points to be on the wrong side of the hyperplane. The cost parameter,
\(C\), dictates the tolerance for points to be on the wrong side of the
margin. A large value of \(C\) allows many points to be on the wrong
side while smaller values of \(C\) have a much lower tolerance for
misclassified points. A kernel, \(K\), maps the classifier into a higher
dimensional space. Hyperplanes are used to classify in the higher
dimensional space, which results in non-linear boundaries in the
original space. The SVM is modeled as:

\[
    f(x) = \beta_0 + \sum_{i \in S} \alpha_i K(x, x_i;\gamma)
\]

where, \(K\) is a kernel with tuning parameter \(\gamma\), \(S\) is the
set of support vectors (points on the boundary of the margin), and
\(\alpha_i\) is computed using \(C\) and the margin. The hyperparameters
for SVM classification are \(C\) and \(\gamma\). Common kernels are
polynomial, radial, and linear. \texttt{tidymodels} and \texttt{caret}
will tune all three types of kernels whereas \texttt{EZtune} provides
automatic tuning only for radial kernels. However, radial kernels work
well in most situations.

Support vector regression (SVR) has an additional tuning parameter,
\(\epsilon\). SVR attempts to find a function, or hyperplane, such that
the deviations between the hyperplane and the responses, \(y_i\), are
less than \(\epsilon\) for each observation \cite{smola2004tutorial}.
The cost represents the number of points that can be further than
\(\epsilon\) away from the hyperplane. Essentially, SVMs try to maximize
the number of points that are on the correct side of the margin and SVR
tries to maximize the number of points that fall within \(\epsilon\) of
the margin. The only mathematical restriction for the hyperparameters
for SVM and SVR is that they are greater than 0.

\hypertarget{gbmintro}{%
\subsection{Boosted trees}\label{gbmintro}}

Boosted trees are part of the family of ensemble methods which combine
many weak learners, or classifiers, into a single, accurate, classifier.
A weak learner typically does not perform well alone, but combining many
weak learners can create a strong classifier \cite{friedman2009elements}. With boosted trees, a weak learning model is constructed
using a regression or classification tree with only a few terminal
nodes. The misclassified points or residuals from this tree are examined
and the information is used to fit a new tree. The model is updated by
adding the new tree to the previously fitted trees. The ensemble is
iteratively updated in this manner and final predictions are made by a
weighted vote of the weak learners.

The primary difference between various boosted tree algorithms is the
method used to learn from misclassified observations at each iteration.
Adaboost fits a small tree to the training data by applying the same
weight to all observations in the training data \cite{adaboost}. The misclassified points are then given greater weight than the
correctly classified points and a new tree is computed. The new
prediction is the sum of the weighted predictions of all of the previous
trees. The process is repeated many times with misclassified points
being given greater weight. A new tree is created using the weighted
data and it is then added to the previous model. The weak learners are
combined using a weighted average approach where the highest weights are
given to the best performing weak learners. This results in an additive
model where the final predictions are the weighted sum of the
predictions made by all of the models in the ensemble \cite{friedman2009elements}.

GBMs are boosted trees that use gradient descent to minimize a loss
function during the learning process \cite{gbmpaper}. The loss
function can be tailored to the problem being solved. We use the mean
squared error (MSE) for regression models and a logarithmic loss for
classification problems as the loss functions for the examples in this
article. A decision tree is used as the initial weak learner. GBMs
recursively fit new trees to the residuals from previous trees and then
combine the predictions from all of the trees to obtain a final
prediction.

Adaboost and GBMs have a nearly identical set of hyperparameters. Both
models require tuning the number of iterations, depth of the trees, and
the shrinkage, which controls how fast the trees learn. GBMs have an
additional hyperparameter which is the minimum number of observations in
a terminal node.

\hypertarget{enintro}{%
\subsection{Elastic net}\label{enintro}}

Elastic net is a linear model that incorporates \(\ell_1\) and
\(\ell_2\) regularization. Regularization reduces variability in the
model with the sacrifice of introducing some bias. The \(\ell_1\)
penalty introduces sparseness into the model. However, using only the
\(\ell_1\) penalty limits the number of variables that can have non-zero
coefficients to the number of observations and prevents group selection
of variables. That is, if a group of variables are correlated, only one
of the variables will typically be selected. Introducing \(\ell_2\)
regularization allows for more non-zero coefficients and encourages
correlated groups of variables to be retained in the model. Elastic net
estimates the coefficients using the following equation:

\[
    \hat{\beta} = \underset{\beta}{\operatorname{argmin}} ||\boldsymbol{y} - \boldsymbol{X}\beta||^2 + \lambda_2 ||\beta||_2^2 + \lambda_1 ||\beta||_1
\]

The parameters \(\lambda_1\) and \(\lambda_2\) control the amount of
\(\ell_1\) and \(\ell_2\) regularization in the model. Ridge regression
is a special case of elastic net where \(\lambda_1 = 0\). The
coefficients shrink toward 0, but none of them will be equal to 0 which
results in the retention of all predictors in the model. Similarly,
lasso is an elastic net model with \(\lambda_2 = 0\) which results in
many coefficients being set to 0. Larger values of \(\lambda_1\) result
in more shrinkage of the coefficients.

Elastic net has two hyperparameters: \(\alpha\) and \(\lambda\). The
parameter \(\alpha\) is the elastic net tuning parameter and it controls
the amount \(\ell_1\) and \(\ell_2\) regularization in the model. It is
defined as \(\alpha = \frac{\lambda_1} {\lambda_1 +\lambda_2}\). Note
that \(\alpha \in [0, 1]\), where \(\alpha = 0\) is the ridge model and
\(\alpha = 1\) is the lasso model. The other tuning parameter,
\(\lambda\), controls the amount of shrinkage that is performed. Larger
values of \(\lambda\) result in more shrinkage. The only mathematical
restriction on \(\lambda\) is that \(\lambda \geq 0\).

\hypertarget{rpackages}{%
\section{Discussion of available R packages}\label{rpackages}}

Several R packages are available that can tune statistical learning
models. Packages such as \texttt{e1071} \cite{e1071} can tune a
single model type. However, we are interested in being able to compare
different model types with a simple interface. Thus, we limit this
discussion to the most commonly used R packages that can tune different
model types: \texttt{caret}, \texttt{tidymodels}, and \texttt{EZtune}.

The \texttt{caret} package \cite{caret} is a powerful package that has
been available in R for many years. \texttt{caret} is able to tune
nearly any model using almost any method. However, this abundant
functionality makes \texttt{caret} time consuming to learn and can be
overwhelming and inaccessible to a non-expert R user. \texttt{caret} is
not used in comparisons in this article because although it is widely
used, we feel that the programming and machine learning knowledge needed
to use it makes \texttt{caret} a poor candidate for comparison with
\texttt{EZtune}.

\texttt{tidymodels} \cite{tidymodels} is a suite of packages that
can automatically tune many supervised learning models with varying
degrees of automation. \texttt{tidymodels} is not as powerful or
versatile as \texttt{caret}, but it is much easier to learn and use.
\texttt{tidymodels} can tune many different model types and allows the
user to tune a model using a grid search or Iterative Bayesian
optimization \cite{iterativebayes}. \texttt{tidymodels} includes functionality
that auto-selects reasonable ranges for the grid search, which is
helpful for the user who is not an expert in hyperparameters. However,
it requires that the user knows what hyperparameters must be tuned and
what R packages are used to construct the different models. Although
\texttt{tidymodels} is much easier to use than \texttt{caret}, it is
still not accessible to a novice R user and takes considerable
understanding of the different models and their hyperparameters to
learn.

\texttt{EZtune} \cite{dissertation} tunes fewer models than \texttt{caret} or
\texttt{tidymodels}, but the user interface is simple and accessible to
those who are novice R users or inexperienced with machine learning
models. Tuning is done by optimizing the hyperparameter space using
either a Hookes-Jeeves algorithm \cite{hookejeeves} or a genetic
algorithm \cite{geneticalgorithm}. \texttt{EZtune} does not require any
knowledge of the hyperparameters or their properties. The interface is
designed to work well within a computational pipeline or R function.

\hypertarget{eztunecomp}{%
\section{Comparison of EZtune with other R packages}\label{eztunecomp}}

This section includes a comparison of \texttt{EZtune} with
\texttt{tidymodels} for tuning SVMs and GBMs. \texttt{tidymodels} does
not tune elastic net models so we include a comparison of
\texttt{EZtune} and \texttt{glmnet} \cite{glmnet} for tuning elastic net. Adaboost is not included in this section
because \texttt{tidymodels} does not tune adaboost. The section is
intended to showcase the strengths and weaknesses of \texttt{tidymodels}
and \texttt{EZtune} and to provide a tutorial on how to use both
packages. Many code snippets are included to demonstrate how to use both
packages for different model types and tuning methods.

Comparisons are made using both classification and regression models
because different models and packages perform differently in each of
these settings. Five datasets have a binary response and are used for
classification and four datasets have a continuous response variable and
are used to compare the regression methods. These datasets were selected
because they are publicly available and have been used in previous
benchmarking studies. A description of the datasets is in Table~\ref{tab:datasets}.

\begin{table}

\caption{List of datasets used to explore hyperparameters.}
\centering
\begin{tabular}[t]{lrrrr}
\toprule
Data sets & N & Variables & Categorical variables & Continuous variables\\
\midrule
Abalone & 4177 & 9 & 1 & 7\\
Boston Housing 2 & 506 & 19 & 1 & 15\\
CO2 & 84 & 5 & 3 & 1\\
Crime & 47 & 14 & 1 & 12\\
Breast Cancer & 699 & 10 & 0 & 9\\
\addlinespace
Pima & 768 & 9 & 0 & 8\\
Sonar & 208 & 61 & 0 & 60\\
Lichen & 840 & 40 & 2 & 31\\
Mullein & 12094 & 32 & 0 & 31\\
\bottomrule
\multicolumn{5}{l}{\rule{0pt}{1em}\textit{Note: }}\\
\multicolumn{5}{l}{\rule{0pt}{1em}Abalone is from the AppliedPredictiveModeling package \cite{apm}.}\\
\multicolumn{5}{l}{\rule{0pt}{1em}Boston Housing 2, Breast cancer, Pima, and Sonar are from the mlbench package \cite{mlbench}.}\\
\multicolumn{5}{l}{\rule{0pt}{1em}CO2 is from the datasets package \cite{R}.}\\
\multicolumn{5}{l}{\rule{0pt}{1em}Crime is from the book Practicing Statistics \cite{kuiper2013practicing}.}\\
\multicolumn{5}{l}{\rule{0pt}{1em}Lichen and Mullein are internal to EZtune \cite{dissertation}.}\\
\end{tabular}
\label{tab:datasets}
\end{table}

Datasets were split into training and test datasets using the
\texttt{rsample} package \cite{rsample} and models were tuned using
the training dataset. Tuned models were verified using the test data
from the split and the results were compared for each method and
dataset. The following code shows how the data were split for all of the
binary classification tests. The same methodology was used for the
regression datasets except that the strata argument is not used in the
initial\_split function.

\begin{Shaded}
\begin{Highlighting}[]
\FunctionTok{library}\NormalTok{(mlbench)}
\FunctionTok{library}\NormalTok{(rsample)}
\FunctionTok{data}\NormalTok{(Sonar)}
\NormalTok{sonar\_split }\OtherTok{\textless{}{-}} \FunctionTok{initial\_split}\NormalTok{(Sonar, }\AttributeTok{strata =}\NormalTok{ Class)}
\NormalTok{sonar\_train }\OtherTok{\textless{}{-}} \FunctionTok{training}\NormalTok{(sonar\_split)}
\NormalTok{sonar\_test }\OtherTok{\textless{}{-}} \FunctionTok{testing}\NormalTok{(sonar\_split)}
\NormalTok{sonar\_folds }\OtherTok{\textless{}{-}} \FunctionTok{vfold\_cv}\NormalTok{(sonar\_train)}
\end{Highlighting}
\end{Shaded}

The model was tuned and accuracy or root mean squared error (RMSE) and
computation time was recorded for ten trials. The mean computation time
and mean accuracy are reported for each dataset and tuning method.
\texttt{EZtune} was tested with both the genetic and the Hooke-Jeeves
algorithms and with 10-fold cross-validation and the fast method for
verification while tuning. The fast method randomly splits the data in
half, trains the model with half of the data, and verifies the model
with the other half \cite{dissertation}. The GBM and SVM comparisons use
\texttt{tidymodels} with both a grid search and Iterative Bayes
optimization. The grid comprised five different values for each
hyperparameter selected by \texttt{tidymodels} and Iterative Bayes was
done using ten iterations. Elastic net was tuned with \texttt{glmnet}
using two different methods specified in Section~\ref{sec:enres}. Each section includes examples of the code used to perform the
computations.

\hypertarget{svmres}{%
\subsection{Results for support vector machines}\label{svmres}}

\texttt{tidymodels} uses the package \texttt{kernlab} \cite{kernlab} and \texttt{EZtune} uses the package
\texttt{e1071} \cite{e1071} as the engine for the SVM
calculations. \texttt{EZtune} only tunes models with a radial kernel,
but \texttt{tidymodels} can tune a model with a linear, polynomial, or
radial kernel. All comparisons were done with radial kernels to ensure
comparability. Cost and \(\gamma\) were both tuned for the binary
classification models and \(\epsilon\) was also tuned for the regression
models.

The following code snippet shows how \texttt{tidymodels} was used to
tune an SVM for the Sonar data using Iterative Bayes optimization. The
model is created by using the svm\_rbf function to specify that the
model is an SVM and to identify the hyperparameters that will be tuned.
This is used in conjunction with set\_engine for specifying the
underlying engine package for SVM computations and set\_mode for
defining the model type. The metrics that will be used to tune the model
are identified with metric\_set. The tuning workflow is then specified
with the workflow, add\_model, and add\_formula functions. Once the
model and the workflow are specified, the parameters from the model, the
workflow, and the set of performance metrics are used by the function
tune\_bayes to tune the SVM using Iterative Bayes. The performance
results are obtained by refitting the tuned model with final\_workflow
and then obtaining the metrics from the test dataset with last\_fit and
collect\_metrics. This workflow provides a great deal of flexibility at
all stages, but it can be challenging to piece together and identify the
inputs to each part.

\begin{Shaded}
\begin{Highlighting}[]
\FunctionTok{library}\NormalTok{(tidymodels)}

\NormalTok{tune\_model }\OtherTok{\textless{}{-}} \FunctionTok{svm\_rbf}\NormalTok{(}\AttributeTok{cost =} \FunctionTok{tune}\NormalTok{(), }\AttributeTok{rbf\_sigma =} \FunctionTok{tune}\NormalTok{()) }\SpecialCharTok{\%\textgreater{}\%}
  \FunctionTok{set\_engine}\NormalTok{(}\StringTok{"kernlab"}\NormalTok{) }\SpecialCharTok{\%\textgreater{}\%}
  \FunctionTok{set\_mode}\NormalTok{(}\StringTok{"classification"}\NormalTok{) }
\NormalTok{mets }\OtherTok{\textless{}{-}} \FunctionTok{metric\_set}\NormalTok{(accuracy, roc\_auc)}

\NormalTok{model\_wf }\OtherTok{\textless{}{-}} \FunctionTok{workflow}\NormalTok{() }\SpecialCharTok{\%\textgreater{}\%}
  \FunctionTok{add\_model}\NormalTok{(tune\_model) }\SpecialCharTok{\%\textgreater{}\%}
  \FunctionTok{add\_formula}\NormalTok{(Class }\SpecialCharTok{\textasciitilde{}}\NormalTok{ .)}

\NormalTok{model\_set }\OtherTok{\textless{}{-}} \FunctionTok{parameters}\NormalTok{(model\_wf)}
\NormalTok{best\_model }\OtherTok{\textless{}{-}}\NormalTok{ model\_wf }\SpecialCharTok{\%\textgreater{}\%}
  \FunctionTok{tune\_bayes}\NormalTok{(}\AttributeTok{resamples =}\NormalTok{ sonar\_folds, }\AttributeTok{param\_info =}\NormalTok{ model\_set,}
             \AttributeTok{initial =} \DecValTok{5}\NormalTok{, }\AttributeTok{iter =} \DecValTok{10}\NormalTok{, }\AttributeTok{metrics =}\NormalTok{ mets) }\SpecialCharTok{\%\textgreater{}\%}
  \FunctionTok{select\_best}\NormalTok{(}\StringTok{"accuracy"}\NormalTok{)}

\NormalTok{results }\OtherTok{\textless{}{-}}\NormalTok{ model\_wf }\SpecialCharTok{\%\textgreater{}\%}
  \FunctionTok{finalize\_workflow}\NormalTok{(best\_model) }\SpecialCharTok{\%\textgreater{}\%}
  \FunctionTok{fit}\NormalTok{(}\AttributeTok{data =}\NormalTok{ sonar\_train) }\SpecialCharTok{\%\textgreater{}\%}
  \FunctionTok{last\_fit}\NormalTok{(sonar\_split, }\AttributeTok{metrics =}\NormalTok{ mets) }\SpecialCharTok{\%\textgreater{}\%}
  \FunctionTok{collect\_metrics}\NormalTok{()}

\FunctionTok{as.data.frame}\NormalTok{(results[, }\FunctionTok{c}\NormalTok{(}\DecValTok{1}\NormalTok{, }\DecValTok{3}\NormalTok{)])}
\end{Highlighting}
\end{Shaded}

The following code snippet demonstrates how \texttt{EZtune} was used to
tune an SVM for the Sonar data using a genetic algorithm and 10-fold
cross-validation. Note that \texttt{EZtune} can tune the SVM with a
single function call to eztune while \texttt{tidymodels} requires
calling ten functions to obtain the tuned model. The method for
obtaining the performance metrics for the test dataset is also far less
complicated and more intuitive for a novice R user than for
\texttt{tidymodels}.

\begin{Shaded}
\begin{Highlighting}[]
\FunctionTok{library}\NormalTok{(EZtune)}

\NormalTok{model }\OtherTok{\textless{}{-}} \FunctionTok{eztune}\NormalTok{(}\AttributeTok{x =} \FunctionTok{subset}\NormalTok{(sonar\_train, }\AttributeTok{select =} \SpecialCharTok{{-}}\NormalTok{Class), }
                \AttributeTok{y =}\NormalTok{ sonar\_train}\SpecialCharTok{$}\NormalTok{Class, }\AttributeTok{method =} \StringTok{"svm"}\NormalTok{, }\AttributeTok{optimizer =} \StringTok{"ga"}\NormalTok{, }
                \AttributeTok{fast =} \ConstantTok{FALSE}\NormalTok{, }\AttributeTok{cross =} \DecValTok{10}\NormalTok{)}

\NormalTok{predictions }\OtherTok{\textless{}{-}} \FunctionTok{predict}\NormalTok{(model, sonar\_test)}
\NormalTok{acc }\OtherTok{\textless{}{-}} \FunctionTok{accuracy\_vec}\NormalTok{(}\AttributeTok{truth =}\NormalTok{ sonar\_test}\SpecialCharTok{$}\NormalTok{Class, }\AttributeTok{estimate =}\NormalTok{ predictions[, }\DecValTok{1}\NormalTok{])}
\NormalTok{auc }\OtherTok{\textless{}{-}} \FunctionTok{roc\_auc\_vec}\NormalTok{(}\AttributeTok{truth =}\NormalTok{ sonar\_test}\SpecialCharTok{$}\NormalTok{Class, }\AttributeTok{estimate =}\NormalTok{ predictions[, }\DecValTok{2}\NormalTok{])}

\FunctionTok{data.frame}\NormalTok{(}\AttributeTok{Accuracy =}\NormalTok{ acc, }\AttributeTok{AUC =}\NormalTok{ auc)}
\end{Highlighting}
\end{Shaded}

The mean accuracy and mean computations times in seconds are shown in
Table~\ref{tab:svmbin} which shows that the best accuracies were obtained from
\texttt{EZtune} for all five datasets. It also shows that the shortest
computation times for all datasets were achieved by \texttt{EZtune} with
the Hooke-Jeeves optimization algorithm and the fast option. Computation
times were faster for all of the \texttt{EZtune} runs than for the
\texttt{tidymodels} with some \texttt{EZtune} runs being as much as 50
to 100 times faster than the \texttt{tidymodels} runs. The exception is
Mullein, the largest dataset, tuned with cross-validation.

\begin{table}

\caption{Mean accuracies and computation times in seconds for ten trials of tuning classification SVMs. The best accuracies and times for each dataset are bolded.}
\centering
\begin{tabular}[t]{l>{}l>{}l>{}l>{}l>{}l>{}l}

\toprule
\multicolumn{1}{c}{ } & \multicolumn{4}{c}{EZtune} & \multicolumn{2}{c}{Tidymodels} \\
\cmidrule(l{3pt}r{3pt}){2-5} \cmidrule(l{3pt}r{3pt}){6-7}
Data & GA CV & GA fast & HJ CV & HJ fast & Grid & IB\\
\midrule
\em{Accuracy} & \em{} & \em{} & \em{} & \em{} & \em{} & \em{}\\
BreastCancer & 0.994 & 0.993 & \textbf{0.996} & 0.993 & 0.965 & 0.965\\
Lichen & \textbf{0.900} & 0.892 & 0.871 & 0.886 & 0.856 & 0.842\\
Mullein & \textbf{0.959} & 0.949 & \textbf{0.959} & 0.957 & 0.884 & 0.916\\
Pima & 0.833 & \textbf{0.847} & 0.827 & 0.822 & 0.763 & 0.737\\
\addlinespace
Sonar & 0.948 & \textbf{0.959} & 0.954 & 0.957 & 0.814 & 0.882\\
\em{Time (seconds)} & \em{} & \em{} & \em{} & \em{} & \em{} & \em{}\\
BreastCancer & 9.32 & 3.47 & 1.35 & \textbf{0.591} & 126 & 88.0\\
Lichen & 59.1 & 15.7 & 14.4 & \textbf{3.74} & 146 & 92.8\\
Mullein & 47,800 & 3,550 & 38,300 & \textbf{1,170} & 7,380 & 3,310\\
\addlinespace
Pima & 38.2 & 9.26 & 5.91 & \textbf{1.38} & 122 & 84.3\\
Sonar & 9.54 & 4.61 & 2.76 & \textbf{1.56} & 111 & 87.7\\
\bottomrule
\end{tabular}
\label{tab:svmbin}
\end{table}

Support vector regression was done on four datasets. The same
methodology was used for the regression model as for the binary
classification model, except that \(\epsilon\) was tuned in addition to
cost and \(\gamma\). The code for regression with \texttt{EZtune} is
identical to that for binary classification because \texttt{EZtune}
automatically make the appropriate adjustments for the type of response
variable. \texttt{tidymodels} requires a slight modification to specify
whether a model is classification or regression. As with the binary
classification SVM trials, the training dataset was used to tune the
model with each method and then the model was verified with the test
dataset.

The following code snippet shows how the Boston Housing dataset was
split for the regression tests.

\begin{Shaded}
\begin{Highlighting}[]
\FunctionTok{library}\NormalTok{(mlbench)}
\FunctionTok{data}\NormalTok{(BostonHousing2)}
\NormalTok{bh }\OtherTok{\textless{}{-}} \FunctionTok{mutate}\NormalTok{(BostonHousing2, }\AttributeTok{lcrim =} \FunctionTok{log}\NormalTok{(crim)) }\SpecialCharTok{\%\textgreater{}\%}
\NormalTok{  dplyr}\SpecialCharTok{::}\FunctionTok{select}\NormalTok{(}\SpecialCharTok{{-}}\NormalTok{town, }\SpecialCharTok{{-}}\NormalTok{medv, }\SpecialCharTok{{-}}\NormalTok{crim)}
\NormalTok{bh\_split }\OtherTok{\textless{}{-}} \FunctionTok{initial\_split}\NormalTok{(bh)}
\NormalTok{bh\_train }\OtherTok{\textless{}{-}} \FunctionTok{training}\NormalTok{(bh\_split)}
\NormalTok{bh\_test }\OtherTok{\textless{}{-}} \FunctionTok{testing}\NormalTok{(bh\_split)}
\NormalTok{bh\_folds }\OtherTok{\textless{}{-}} \FunctionTok{vfold\_cv}\NormalTok{(bh\_train)}
\end{Highlighting}
\end{Shaded}

The following code snippet demonstrates how an SVM was tuned for the
Boston Housing data using \texttt{tidymodels} with Iterative Bayes
optimization. The workflow is similar to the one used for the SVM for
binary classification. The primary differences are that the model is
specified as a regression model, \(\epsilon\) is added as a
hyperparameter, and the metrics used to verify the model are RMSE and
mean absolute error.

\begin{Shaded}
\begin{Highlighting}[]
\NormalTok{tune\_model }\OtherTok{\textless{}{-}} \FunctionTok{svm\_rbf}\NormalTok{(}\AttributeTok{cost =} \FunctionTok{tune}\NormalTok{(), }\AttributeTok{rbf\_sigma =} \FunctionTok{tune}\NormalTok{(), }\AttributeTok{margin =} \FunctionTok{tune}\NormalTok{()) }\SpecialCharTok{\%\textgreater{}\%}
  \FunctionTok{set\_engine}\NormalTok{(}\StringTok{"kernlab"}\NormalTok{) }\SpecialCharTok{\%\textgreater{}\%}
  \FunctionTok{set\_mode}\NormalTok{(}\StringTok{"regression"}\NormalTok{) }
\NormalTok{mets }\OtherTok{\textless{}{-}} \FunctionTok{metric\_set}\NormalTok{(rmse, mae)}

\NormalTok{model\_wf }\OtherTok{\textless{}{-}} \FunctionTok{workflow}\NormalTok{() }\SpecialCharTok{\%\textgreater{}\%}
  \FunctionTok{add\_model}\NormalTok{(tune\_model) }\SpecialCharTok{\%\textgreater{}\%}
  \FunctionTok{add\_formula}\NormalTok{(cmedv }\SpecialCharTok{\textasciitilde{}}\NormalTok{ .) }

\NormalTok{model\_set }\OtherTok{\textless{}{-}} \FunctionTok{parameters}\NormalTok{(model\_wf)}
\NormalTok{best\_model }\OtherTok{\textless{}{-}}\NormalTok{ model\_wf }\SpecialCharTok{\%\textgreater{}\%}
  \FunctionTok{tune\_bayes}\NormalTok{(}\AttributeTok{resamples =}\NormalTok{ bh\_folds, }\AttributeTok{param\_info =}\NormalTok{ model\_set,}
             \AttributeTok{initial =} \DecValTok{5}\NormalTok{, }\AttributeTok{iter =} \DecValTok{10}\NormalTok{, }\AttributeTok{metrics =}\NormalTok{ mets) }\SpecialCharTok{\%\textgreater{}\%}
  \FunctionTok{select\_best}\NormalTok{(}\StringTok{"rmse"}\NormalTok{)}

\NormalTok{results }\OtherTok{\textless{}{-}}\NormalTok{ model\_wf }\SpecialCharTok{\%\textgreater{}\%}
  \FunctionTok{finalize\_workflow}\NormalTok{(best\_model) }\SpecialCharTok{\%\textgreater{}\%}
  \FunctionTok{fit}\NormalTok{(}\AttributeTok{data =}\NormalTok{ bh\_train) }\SpecialCharTok{\%\textgreater{}\%}
  \FunctionTok{last\_fit}\NormalTok{(bh\_split, }\AttributeTok{metrics =}\NormalTok{ mets) }\SpecialCharTok{\%\textgreater{}\%}
  \FunctionTok{collect\_metrics}\NormalTok{()}

\FunctionTok{as.data.frame}\NormalTok{(results[, }\FunctionTok{c}\NormalTok{(}\DecValTok{1}\NormalTok{, }\DecValTok{3}\NormalTok{)])}
\end{Highlighting}
\end{Shaded}

The following code snippet demonstrates how an SVM was tuned for the
Boston Housing data with a genetic algorithm and 10-fold
cross-validation using \texttt{EZtune}. Note that the syntax for using
eztune is the same as for the binary classification SVM. This is because
eztune uses the response variable to determine if the model is a
classification model or a regression model and then adjusts the
hyperparameters, tuning regions, and verification metrics accordingly.

\begin{Shaded}
\begin{Highlighting}[]
\NormalTok{model }\OtherTok{\textless{}{-}} \FunctionTok{eztune}\NormalTok{(}\AttributeTok{x =} \FunctionTok{subset}\NormalTok{(bh\_train, }\AttributeTok{select =} \SpecialCharTok{{-}}\NormalTok{cmedv), }\AttributeTok{y =}\NormalTok{ bh\_train}\SpecialCharTok{$}\NormalTok{cmedv,}
                \AttributeTok{method =} \StringTok{"svm"}\NormalTok{, }\AttributeTok{optimizer =} \StringTok{"ga"}\NormalTok{, }\AttributeTok{fast =} \ConstantTok{FALSE}\NormalTok{, }\AttributeTok{cross =} \DecValTok{10}\NormalTok{)}

\NormalTok{predictions }\OtherTok{\textless{}{-}} \FunctionTok{predict}\NormalTok{(model, bh\_test)}
\NormalTok{rmse.ez }\OtherTok{\textless{}{-}} \FunctionTok{rmse\_vec}\NormalTok{(}\AttributeTok{truth =}\NormalTok{ bh\_test}\SpecialCharTok{$}\NormalTok{cmedv, }\AttributeTok{estimate =}\NormalTok{ predictions)}
\NormalTok{mae.ez }\OtherTok{\textless{}{-}} \FunctionTok{mae\_vec}\NormalTok{(}\AttributeTok{truth =}\NormalTok{ bh\_test}\SpecialCharTok{$}\NormalTok{cmedv, }\AttributeTok{estimate =}\NormalTok{ predictions)}
\FunctionTok{data.frame}\NormalTok{(}\AttributeTok{RMSE =}\NormalTok{ rmse.ez, }\AttributeTok{MAE =}\NormalTok{ mae.ez)}
\end{Highlighting}
\end{Shaded}

The RMSE was computed for ten runs of each model type and the mean RMSE
is listed for each method and dataset in Table~\ref{tab:svmreg} along with the mean
computation time for each run. The table shows that the RMSEs for each
method are similar, but all of the smallest RMSEs were obtained with
\texttt{EZtune}. The shortest computation times were achieved with
\texttt{EZtune} using the Hooke-Jeeves algorithm and fast option. The
longest computation time was seen with the Abalone data for
\texttt{EZtune} with the genetic algorithm and cross-validation. This
mirrors what was seen with the binary classification results in Table~\ref{tab:svmbin}
which also showed that the genetic algorithm with cross-validation on
large datasets is computationally slower than the other \texttt{EZtune}
and \texttt{tidymodels} options. The accuracies and RMSEs for the
cross-validated genetic algorithm are not better than the other options
which implies it may not be worth the long computation time for larger
datasets.

\begin{table}

\caption{Mean RMSEs and computation times in seconds for ten trials of tuning regression SVMs. The best results for each dataset are bolded.}
\centering
\begin{tabular}[t]{l>{}l>{}l>{}l>{}l>{}l>{}l}
\toprule
\multicolumn{1}{c}{ } & \multicolumn{4}{c}{EZtune} & \multicolumn{2}{c}{Tidymodels} \\
\cmidrule(l{3pt}r{3pt}){2-5} \cmidrule(l{3pt}r{3pt}){6-7}
Data & GA CV & GA fast & HJ CV & HJ fast & Grid & IB\\
\midrule
\em{RMSE} & \em{} & \em{} & \em{} & \em{} & \em{} & \em{}\\
Abalone & 2.16 & 2.15 & \textbf{2.09} & 2.11 & 2.11 & 2.13\\
BostonHousing & \textbf{2.82} & 3.12 & 3.50 & 2.94 & 3.09 & 2.89\\
CO2 & 4.20 & \textbf{3.80} & 4.44 & 4.31 & 4.28 & 4.79\\
Crime & 26.7 & 28.9 & \textbf{24.6} & 26.8 & 30.2 & 28.0\\
\addlinespace
\em{Time (seconds)} & \em{} & \em{} & \em{} & \em{} & \em{} & \em{}\\
Abalone & 8,410 & 272 & 309 & \textbf{26.5} & 1,980 & 327\\
BostonHousing & 91.5 & 4.75 & 41.0 & \textbf{1.56} & 426 & 91.1\\
CO2 & 6.99 & 2.06 & 1.49 & \textbf{0.442} & 414 & 137\\
Crime & 1.93 & 1.87 & 0.573 & \textbf{0.436} & 447 & 107\\
\bottomrule
\end{tabular}
\label{tab:svmreg}
\end{table}

\hypertarget{gbmres}{%
\subsection{Results for gradient boosting machines}\label{gbmres}}

\texttt{tidymodels} uses the package \texttt{xgboost} \cite{xgboost}
and \texttt{EZtune} uses the package \texttt{gbm} \cite{gbm} as the engine for GBM. All four tuning parameters for GBMs were
tuned with \texttt{tidymodels} and \texttt{EZtune}. As with SVMs,
\texttt{tidymodels} was run with both a grid search and an Iterative
Bayes algorithm using the same criteria for grid size and iterations as
with SVMs. \texttt{EZtune} was run using the same criteria that was used
for the SVM iterations.

The following code snippet shows the code used to tune a GBM for the
Sonar data with \texttt{tidymodels} using a grid search.

\begin{Shaded}
\begin{Highlighting}[]
\NormalTok{tune\_model }\OtherTok{\textless{}{-}} \FunctionTok{boost\_tree}\NormalTok{(}\AttributeTok{trees =} \FunctionTok{tune}\NormalTok{(), }\AttributeTok{tree\_depth =} \FunctionTok{tune}\NormalTok{(),}
                         \AttributeTok{learn\_rate =} \FunctionTok{tune}\NormalTok{(),}
                         \AttributeTok{min\_n =} \FunctionTok{tune}\NormalTok{()) }\SpecialCharTok{\%\textgreater{}\%}
  \FunctionTok{set\_engine}\NormalTok{(}\StringTok{"xgboost"}\NormalTok{) }\SpecialCharTok{\%\textgreater{}\%}
  \FunctionTok{set\_mode}\NormalTok{(}\StringTok{"classification"}\NormalTok{) }
\NormalTok{mets }\OtherTok{\textless{}{-}} \FunctionTok{metric\_set}\NormalTok{(accuracy, roc\_auc)}

\NormalTok{model\_wf }\OtherTok{\textless{}{-}} \FunctionTok{workflow}\NormalTok{() }\SpecialCharTok{\%\textgreater{}\%}
  \FunctionTok{add\_model}\NormalTok{(tune\_model) }\SpecialCharTok{\%\textgreater{}\%}
  \FunctionTok{add\_formula}\NormalTok{(Class }\SpecialCharTok{\textasciitilde{}}\NormalTok{ .)}

\NormalTok{best\_model }\OtherTok{\textless{}{-}}\NormalTok{ model\_wf }\SpecialCharTok{\%\textgreater{}\%}
  \FunctionTok{tune\_grid}\NormalTok{(}\AttributeTok{resamples =}\NormalTok{ sonar\_folds, }\AttributeTok{grid =} \DecValTok{5}\SpecialCharTok{\^{}}\DecValTok{4}\NormalTok{, }\AttributeTok{metrics =}\NormalTok{ mets) }\SpecialCharTok{\%\textgreater{}\%} 
  \FunctionTok{select\_best}\NormalTok{(}\StringTok{"accuracy"}\NormalTok{)}

\NormalTok{results }\OtherTok{\textless{}{-}}\NormalTok{ model\_wf }\SpecialCharTok{\%\textgreater{}\%}
  \FunctionTok{finalize\_workflow}\NormalTok{(best\_model) }\SpecialCharTok{\%\textgreater{}\%}
  \FunctionTok{fit}\NormalTok{(}\AttributeTok{data =}\NormalTok{ sonar\_train) }\SpecialCharTok{\%\textgreater{}\%}
  \FunctionTok{last\_fit}\NormalTok{(sonar\_split, }\AttributeTok{metrics =}\NormalTok{ mets) }\SpecialCharTok{\%\textgreater{}\%}
  \FunctionTok{collect\_metrics}\NormalTok{()}
\FunctionTok{as.data.frame}\NormalTok{(results[, }\FunctionTok{c}\NormalTok{(}\DecValTok{1}\NormalTok{, }\DecValTok{3}\NormalTok{)])}
\end{Highlighting}
\end{Shaded}

The following code snippet demonstrates how a GBM was tuned for the
Sonar data using \texttt{EZtune} with Hooke-Jeeves and the fast option.

\begin{Shaded}
\begin{Highlighting}[]
\NormalTok{model }\OtherTok{\textless{}{-}} \FunctionTok{eztune}\NormalTok{(}\AttributeTok{x =} \FunctionTok{subset}\NormalTok{(sonar\_train, }\AttributeTok{select =} \SpecialCharTok{{-}}\NormalTok{Class), }
                \AttributeTok{y =}\NormalTok{ sonar\_train}\SpecialCharTok{$}\NormalTok{Class, }\AttributeTok{method =} \StringTok{"gbm"}\NormalTok{, }\AttributeTok{optimizer =} \StringTok{"hjn"}\NormalTok{, }
                \AttributeTok{fast =} \FloatTok{0.5}\NormalTok{)}

\NormalTok{predictions }\OtherTok{\textless{}{-}} \FunctionTok{predict}\NormalTok{(model, sonar\_test)}
\NormalTok{acc }\OtherTok{\textless{}{-}} \FunctionTok{accuracy\_vec}\NormalTok{(}\AttributeTok{truth =}\NormalTok{ sonar\_test}\SpecialCharTok{$}\NormalTok{Class, }\AttributeTok{estimate =}\NormalTok{ predictions[, }\DecValTok{1}\NormalTok{])}
\NormalTok{auc }\OtherTok{\textless{}{-}} \FunctionTok{roc\_auc\_vec}\NormalTok{(}\AttributeTok{truth =}\NormalTok{ sonar\_test}\SpecialCharTok{$}\NormalTok{Class, }\AttributeTok{estimate =}\NormalTok{ predictions[, }\DecValTok{2}\NormalTok{])}
\FunctionTok{data.frame}\NormalTok{(}\AttributeTok{Accuracy =}\NormalTok{ acc, }\AttributeTok{AUC =}\NormalTok{ auc)}
\end{Highlighting}
\end{Shaded}

Table~\ref{tab:gbmbin} shows the mean accuracies and the mean computation times for the
ten trials. The table shows that the accuracies for \texttt{EZtune} are
notably higher than those for \texttt{tidymodels} with the difference
being about 3 percentage points for the Breast Cancer data and as large
as 14 percentage points for the Sonar data. The shortest computation
times were seen for \texttt{EZtune} with the Hooke-Jeeves algorithm and
the fast option for all of the datasets with computation times that were
approximately 10 times or more faster than those for the
\texttt{tidymodels} Iterative Bayes option. The accuracies for the
Hooke-Jeeves fast option were also similar to the optimal accuracy
obtained for all of the datasets. The grid search option for
\texttt{tidymodels} was much slower than the other models. This is
because five options were tested for each hyperparameter. The grid for
classification with GBM had 625 tests instead of the 25 needed to tune
an SVM for binary classification. With the exception of the Sonar data,
Iterative Bayes worked nearly as well as the grid search for
\texttt{tidymodels}.

\begin{table}

\caption{Mean accuracies and computation times in seconds for ten trials of tuning classification GBMs. The best results for each dataset are bolded.}
\centering
\begin{tabular}[t]{l>{}l>{}l>{}l>{}l>{}l>{}l}
\toprule
\multicolumn{1}{c}{ } & \multicolumn{4}{c}{EZtune} & \multicolumn{2}{c}{Tidymodels} \\
\cmidrule(l{3pt}r{3pt}){2-5} \cmidrule(l{3pt}r{3pt}){6-7}
Data & GA CV & GA fast & HJ CV & HJ fast & Grid & IB\\
\midrule
\em{Accuracy} & \em{} & \em{} & \em{} & \em{} & \em{} & \em{}\\
BreastCancer & 0.991 & 0.992 & 0.994 & \textbf{0.995} & 0.965 & 0.965\\
Lichen & 0.895 & \textbf{0.898} & 0.891 & 0.893 & 0.844 & 0.852\\
Mullein & \textbf{0.970} & \textbf{0.970} & 0.967 & 0.966 & 0.929 & 0.922\\
Pima & \textbf{0.833} & 0.823 & 0.806 & 0.815 & 0.742 & 0.740\\
\addlinespace
Sonar & 0.924 & \textbf{0.935} & 0.934 & 0.904 & 0.863 & 0.794\\
\em{Time (seconds)} & \em{} & \em{} & \em{} & \em{} & \em{} & \em{}\\
BreastCancer & 1,440 & 79.0 & 199 & \textbf{13.9} & 5,540 & 196\\
Lichen & 4,130 & 369 & 853 & \textbf{50.3} & 11,200 & 433\\
Mullein & 149,000 & 7,770 & 24,600 & \textbf{1,160} & 194,000 & 7,970\\
\addlinespace
Pima & 935 & 66.5 & 210 & \textbf{13.6} & 5,050 & 208\\
Sonar & 1,730 & 79.5 & 306 & \textbf{17.8} & 6,170 & 200\\
\bottomrule
\end{tabular}
\label{tab:gbmbin}
\end{table}

The following code demonstrates how \texttt{tidymodels} was used to tune
a GBM on the Boston Housing data using a grid search.

\begin{Shaded}
\begin{Highlighting}[]
\NormalTok{tune\_model }\OtherTok{\textless{}{-}} \FunctionTok{boost\_tree}\NormalTok{(}\AttributeTok{trees =} \FunctionTok{tune}\NormalTok{(), }\AttributeTok{tree\_depth =} \FunctionTok{tune}\NormalTok{(),}
                         \AttributeTok{learn\_rate =} \FunctionTok{tune}\NormalTok{(), }\AttributeTok{min\_n =} \FunctionTok{tune}\NormalTok{()) }\SpecialCharTok{\%\textgreater{}\%}
  \FunctionTok{set\_engine}\NormalTok{(}\StringTok{"xgboost"}\NormalTok{) }\SpecialCharTok{\%\textgreater{}\%}
  \FunctionTok{set\_mode}\NormalTok{(}\StringTok{"regression"}\NormalTok{) }
\NormalTok{mets }\OtherTok{\textless{}{-}} \FunctionTok{metric\_set}\NormalTok{(rmse, mae)}

\NormalTok{model\_wf }\OtherTok{\textless{}{-}} \FunctionTok{workflow}\NormalTok{() }\SpecialCharTok{\%\textgreater{}\%}
  \FunctionTok{add\_model}\NormalTok{(tune\_model) }\SpecialCharTok{\%\textgreater{}\%}
  \FunctionTok{add\_formula}\NormalTok{(cmedv }\SpecialCharTok{\textasciitilde{}}\NormalTok{ .) }

\NormalTok{best\_model }\OtherTok{\textless{}{-}}\NormalTok{ model\_wf }\SpecialCharTok{\%\textgreater{}\%}
  \FunctionTok{tune\_grid}\NormalTok{(}\AttributeTok{resamples =}\NormalTok{ bh\_folds, }\AttributeTok{grid =} \DecValTok{5}\SpecialCharTok{\^{}}\DecValTok{4}\NormalTok{, }\AttributeTok{metrics =}\NormalTok{ mets) }\SpecialCharTok{\%\textgreater{}\%} 
  \FunctionTok{select\_best}\NormalTok{(}\StringTok{"rmse"}\NormalTok{)}

\NormalTok{results }\OtherTok{\textless{}{-}}\NormalTok{ model\_wf }\SpecialCharTok{\%\textgreater{}\%}
  \FunctionTok{finalize\_workflow}\NormalTok{(best\_model) }\SpecialCharTok{\%\textgreater{}\%}
  \FunctionTok{fit}\NormalTok{(}\AttributeTok{data =}\NormalTok{ bh\_train) }\SpecialCharTok{\%\textgreater{}\%}
  \FunctionTok{last\_fit}\NormalTok{(bh\_split, }\AttributeTok{metrics =}\NormalTok{ mets) }\SpecialCharTok{\%\textgreater{}\%}
  \FunctionTok{collect\_metrics}\NormalTok{()}
\FunctionTok{as.data.frame}\NormalTok{(results[, }\FunctionTok{c}\NormalTok{(}\DecValTok{1}\NormalTok{, }\DecValTok{3}\NormalTok{)])}
\end{Highlighting}
\end{Shaded}

The following code snippet shows how to tune a GBM for the Boston
Housing data using \texttt{EZtune} with Hooke-Jeeves and the fast
option.

\begin{Shaded}
\begin{Highlighting}[]
\NormalTok{model }\OtherTok{\textless{}{-}} \FunctionTok{eztune}\NormalTok{(}\AttributeTok{x =} \FunctionTok{subset}\NormalTok{(bh\_train, }\AttributeTok{select =} \SpecialCharTok{{-}}\NormalTok{cmedv), }\AttributeTok{y =}\NormalTok{ bh\_train}\SpecialCharTok{$}\NormalTok{cmedv,}
                \AttributeTok{method =} \StringTok{"gbm"}\NormalTok{, }\AttributeTok{optimizer =} \StringTok{"hjn"}\NormalTok{, }\AttributeTok{fast =} \FloatTok{0.5}\NormalTok{)}


\NormalTok{predictions }\OtherTok{\textless{}{-}} \FunctionTok{predict}\NormalTok{(model, bh\_test)}
\NormalTok{rmse.ez }\OtherTok{\textless{}{-}} \FunctionTok{rmse\_vec}\NormalTok{(}\AttributeTok{truth =}\NormalTok{ bh\_test}\SpecialCharTok{$}\NormalTok{cmedv, }\AttributeTok{estimate =}\NormalTok{ predictions)}
\NormalTok{mae.ez }\OtherTok{\textless{}{-}} \FunctionTok{mae\_vec}\NormalTok{(}\AttributeTok{truth =}\NormalTok{ bh\_test}\SpecialCharTok{$}\NormalTok{cmedv, }\AttributeTok{estimate =}\NormalTok{ predictions)}
\FunctionTok{data.frame}\NormalTok{(}\AttributeTok{RMSE =}\NormalTok{ rmse.ez, }\AttributeTok{MAE =}\NormalTok{ mae.ez)}
\end{Highlighting}
\end{Shaded}

Table~\ref{tab:gbmreg} shows the mean RMSEs and the mean computation times for the
regression trials. As with binary classification, the grid search with
\texttt{tidymodels} is substantially slower than the other options
without meaningful improvements in RMSE. The \texttt{EZtune} fast
computations have much shorter computation times than the other methods,
with Hooke-Jeeves having the shortest computation times. The best RMSE
results for three of the four datasets were achieved by \texttt{EZtune}.

\begin{table}

\caption{Mean RMSEs and computation times in seconds for ten trials of tuning regression GBMs. The the best results for each dataset are bolded.}
\centering
\begin{tabular}[t]{l>{}l>{}l>{}l>{}l>{}l>{}l}
\toprule
\multicolumn{1}{c}{ } & \multicolumn{4}{c}{EZtune} & \multicolumn{2}{c}{Tidymodels} \\
\cmidrule(l{3pt}r{3pt}){2-5} \cmidrule(l{3pt}r{3pt}){6-7}
Data & GA CV & GA fast & HJ CV & HJ fast & Grid & IB\\
\midrule
\em{RMSE} & \em{} & \em{} & \em{} & \em{} & \em{} & \em{}\\
Abalone & 2.18 & 2.14 & 2.17 & 2.16 & \textbf{2.13} & 2.15\\
BostonHousing & 2.63 & 2.79 & 2.97 & \textbf{2.48} & 2.92 & 3.00\\
CO2 & 2.60 & 2.80 & \textbf{2.48} & 2.71 & 2.59 & 2.55\\
Crime & 25.8 & 31.2 & 27.5 & \textbf{22.6} & 24.1 & 31.2\\
\addlinespace
\em{Time (seconds)} & \em{} & \em{} & \em{} & \em{} & \em{} & \em{}\\
Abalone & 8,110 & 523 & 4,180 & \textbf{289} & 32,800 & 675\\
BostonHousing & 3,180 & 171 & 1,480 & \textbf{64.0} & 6,840 & 288\\
CO2 & 98.6 & 6.54 & 49.9 & \textbf{3.35} & 3,380 & 176\\
Crime & 81.3 & 2.84 & 41.2 & \textbf{1.20} & 3,600 & 128\\
\bottomrule
\end{tabular}
\label{tab:gbmreg}
\end{table}

\hypertarget{sec:enres}{%
\subsection{Results for elastic net}\label{sec:enres}}

\texttt{glmnet} \cite{glmnet} can be used
to tune elastic net, but it will not tune both \(\lambda\) and
\(\alpha\) simulataneously. Automatic tuning with \texttt{EZtune} is
compared to a common tuning method using \texttt{glmnet}. The
\texttt{glmnet} method is as follows:

\begin{enumerate}
\def\labelenumi{\arabic{enumi}.}
\tightlist
\item
  For each \(\alpha\) in (0, 0.1, 0.2, \ldots, 0.9, 1.0) do the
  following:
\item
  Use cv.glmnet to find the \(\lambda\) that achieves the best accuracy
  or RMSE (min-\(\lambda\)) and the \(\lambda\) that produces the the
  error that is within one standard error of the minimum (1-SE).
\item
  Select the \(\alpha\) and \(\lambda\) combination that produces the
  model with the best accuracy or RMSE. Do this for each \(\lambda\)
  type (min-\(\lambda\) and 1-SE).
\end{enumerate}

As with the previous comparisons, the \texttt{EZtune} and the
\texttt{glmnet} models are tuned using a trial dataset and then verified
using a test dataset. Note that \texttt{EZtune} uses \texttt{glmnet} to
simultaneously tune \(\lambda\) and \(\alpha\) but it uses a
Hooke-Jeeves or genetic algorithm to search through the hyperparameter
space.

The following code snippet demonstrates how elastic net was tuned on the
Sonar data using \texttt{glmnet}. Note that \texttt{glmnet} is
particular about how the data are formatted for use in the glmnet and
cv.glmnet functions. The explanatory variables must be a matrix which
means factor or character variables cannot be directly used in the
functions. \texttt{EZtune} is liberal with the way data are passed to
the function eztune. It can handle both data.frame and matrix objects
and can handle both character and factor variables directly.

\begin{Shaded}
\begin{Highlighting}[]
\FunctionTok{library}\NormalTok{(glmnet)}

\NormalTok{foldid }\OtherTok{\textless{}{-}} \FunctionTok{sample}\NormalTok{(}\DecValTok{1}\SpecialCharTok{:}\DecValTok{10}\NormalTok{, }\AttributeTok{size =} \FunctionTok{nrow}\NormalTok{(sonar\_train), }\AttributeTok{replace =} \ConstantTok{TRUE}\NormalTok{)}
\NormalTok{alpha }\OtherTok{\textless{}{-}} \FunctionTok{seq}\NormalTok{(}\DecValTok{0}\NormalTok{, }\DecValTok{1}\NormalTok{, }\FloatTok{0.1}\NormalTok{)}
\NormalTok{alpha\_data }\OtherTok{\textless{}{-}} \FunctionTok{data.frame}\NormalTok{(}\AttributeTok{alpha =}\NormalTok{ alpha, }\AttributeTok{lambda =} \ConstantTok{NA}\NormalTok{, }\AttributeTok{loss =} \ConstantTok{NA}\NormalTok{)}
\NormalTok{model\_cv }\OtherTok{\textless{}{-}} \ConstantTok{NULL}

\ControlFlowTok{for}\NormalTok{ (i }\ControlFlowTok{in} \DecValTok{1}\SpecialCharTok{:}\FunctionTok{length}\NormalTok{(alpha)) \{}
\NormalTok{  model\_cv[[i]] }\OtherTok{\textless{}{-}} \FunctionTok{cv.glmnet}\NormalTok{(}\AttributeTok{x =} \FunctionTok{as.matrix}\NormalTok{(}\FunctionTok{subset}\NormalTok{(sonar\_train, }\AttributeTok{select =} \SpecialCharTok{{-}}\NormalTok{Class)), }
                             \AttributeTok{y =}\NormalTok{ sonar\_train}\SpecialCharTok{$}\NormalTok{Class, }\AttributeTok{family =} \StringTok{"binomial"}\NormalTok{, }
                             \AttributeTok{type.measure =} \StringTok{"class"}\NormalTok{)}
\NormalTok{  alpha\_data[i, }\SpecialCharTok{{-}}\DecValTok{1}\NormalTok{] }\OtherTok{\textless{}{-}} \FunctionTok{c}\NormalTok{(model\_cv[[i]]}\SpecialCharTok{$}\NormalTok{lambda}\FloatTok{.1}\NormalTok{se,}
\NormalTok{                         model\_cv[[i]]}\SpecialCharTok{$}\NormalTok{cvm[model\_cv[[i]]}\SpecialCharTok{$}\NormalTok{lambda }\SpecialCharTok{==} 
\NormalTok{                                             model\_cv[[i]]}\SpecialCharTok{$}\NormalTok{lambda}\FloatTok{.1}\NormalTok{se])}
\NormalTok{\}}

\NormalTok{model }\OtherTok{\textless{}{-}} \FunctionTok{glmnet}\NormalTok{(}\AttributeTok{x =} \FunctionTok{as.matrix}\NormalTok{(}\FunctionTok{subset}\NormalTok{(sonar\_train, }\AttributeTok{select =} \SpecialCharTok{{-}}\NormalTok{Class)), }
                \AttributeTok{y =}\NormalTok{ sonar\_train}\SpecialCharTok{$}\NormalTok{Class, }\AttributeTok{family =} \StringTok{"binomial"}\NormalTok{,}
                \AttributeTok{lambda =}\NormalTok{ alpha\_data}\SpecialCharTok{$}\NormalTok{lambda[alpha\_data}\SpecialCharTok{$}\NormalTok{loss }\SpecialCharTok{==} 
                                             \FunctionTok{min}\NormalTok{(alpha\_data}\SpecialCharTok{$}\NormalTok{loss)][}\DecValTok{1}\NormalTok{],}
                \AttributeTok{alpha =}\NormalTok{ alpha\_data}\SpecialCharTok{$}\NormalTok{alpha[alpha\_data}\SpecialCharTok{$}\NormalTok{loss }\SpecialCharTok{==} 
                                           \FunctionTok{min}\NormalTok{(alpha\_data}\SpecialCharTok{$}\NormalTok{loss)][}\DecValTok{1}\NormalTok{],}
                \AttributeTok{type.measure =} \StringTok{"class"}\NormalTok{)}

\NormalTok{sonar\_test\_truth }\OtherTok{\textless{}{-}} \FunctionTok{as.factor}\NormalTok{(}\FunctionTok{as.numeric}\NormalTok{(sonar\_test}\SpecialCharTok{$}\NormalTok{Class) }\SpecialCharTok{{-}} \DecValTok{1}\NormalTok{)}
\NormalTok{result }\OtherTok{\textless{}{-}} \FunctionTok{predict}\NormalTok{(model, }\FunctionTok{as.matrix}\NormalTok{(}\FunctionTok{subset}\NormalTok{(sonar\_test, }\AttributeTok{select =} \SpecialCharTok{{-}}\NormalTok{Class)), }
                  \AttributeTok{type =} \StringTok{"response"}\NormalTok{)}
\NormalTok{result.r }\OtherTok{\textless{}{-}} \FunctionTok{as.factor}\NormalTok{(}\FunctionTok{round}\NormalTok{(result))}
\NormalTok{acc }\OtherTok{\textless{}{-}} \FunctionTok{accuracy\_vec}\NormalTok{(}\AttributeTok{truth =}\NormalTok{ sonar\_test\_truth, }\AttributeTok{estimate =}\NormalTok{ result.r)}
\NormalTok{auc }\OtherTok{\textless{}{-}} \FunctionTok{roc\_auc\_vec}\NormalTok{(}\AttributeTok{truth =}\NormalTok{ sonar\_test\_truth, }\AttributeTok{estimate =}\NormalTok{ result[, }\DecValTok{1}\NormalTok{],}
                   \AttributeTok{event\_level =} \StringTok{"second"}\NormalTok{)}

\FunctionTok{data.frame}\NormalTok{(}\AttributeTok{Accuracy =}\NormalTok{ acc, }\AttributeTok{AUC =}\NormalTok{ auc)}
\end{Highlighting}
\end{Shaded}

The following code snippet shows how \texttt{EZtune} was used to tune
and elastic net model using the Hooke-Jeeves algorithm and 10-fold
cross-validation. Note that it is much easier to tune an elastic net
model with \texttt{EZtune} than with \texttt{glmnet}.

\begin{Shaded}
\begin{Highlighting}[]
\NormalTok{model }\OtherTok{\textless{}{-}} \FunctionTok{eztune}\NormalTok{(}\AttributeTok{x =} \FunctionTok{subset}\NormalTok{(sonar\_train, }\AttributeTok{select =} \SpecialCharTok{{-}}\NormalTok{Class), }
                \AttributeTok{y =}\NormalTok{ sonar\_train}\SpecialCharTok{$}\NormalTok{Class, }\AttributeTok{method =} \StringTok{"en"}\NormalTok{, }\AttributeTok{optimizer =} \StringTok{"hjn"}\NormalTok{, }
                \AttributeTok{fast =} \ConstantTok{FALSE}\NormalTok{, }\AttributeTok{cross =} \DecValTok{10}\NormalTok{)}

\NormalTok{predictions }\OtherTok{\textless{}{-}} \FunctionTok{predict}\NormalTok{(model, sonar\_test)}
\NormalTok{acc }\OtherTok{\textless{}{-}} \FunctionTok{accuracy\_vec}\NormalTok{(}\AttributeTok{truth =}\NormalTok{ sonar\_test}\SpecialCharTok{$}\NormalTok{Class, }\AttributeTok{estimate =}\NormalTok{ predictions[, }\DecValTok{1}\NormalTok{])}
\NormalTok{auc }\OtherTok{\textless{}{-}} \FunctionTok{roc\_auc\_vec}\NormalTok{(}\AttributeTok{truth =}\NormalTok{ sonar\_test}\SpecialCharTok{$}\NormalTok{Class, }\AttributeTok{estimate =}\NormalTok{ predictions[, }\DecValTok{2}\NormalTok{])}

\FunctionTok{data.frame}\NormalTok{(}\AttributeTok{Accuracy =}\NormalTok{ acc, }\AttributeTok{AUC =}\NormalTok{ auc)}
\end{Highlighting}
\end{Shaded}

Table~\ref{tab:enbin} shows the mean accuracies and mean computation times for all ten
trials. The table shows that no one method produced the best accuracy
for all or most of the datasets and that the accuracies were similar.
The computation times were much faster for \texttt{EZtune} with the
Hooke-Jeeves optimizer and fast option than for the other options. This
was also the best option in terms of accuracy for two of the datasets.

\begin{table}

\caption{Mean accuracy and computation times in seconds for ten trials of tuning classification elastic net models. The best results for each dataset are bolded.}
\centering
\begin{tabular}[t]{l>{}l>{}l>{}l>{}l>{}l>{}l}
\toprule
\multicolumn{1}{c}{ } & \multicolumn{4}{c}{EZtune} & \multicolumn{2}{c}{Glmnet} \\
\cmidrule(l{3pt}r{3pt}){2-5} \cmidrule(l{3pt}r{3pt}){6-7}
Data & GA CV & GA fast & HJ CV & HJ fast & 1-SE & Min\\
\midrule
\em{Accuracy} & \em{} & \em{} & \em{} & \em{} & \em{} & \em{}\\
BreastCancer & 0.959 & \textbf{0.971} & 0.962 & \textbf{0.971} & 0.959 & 0.965\\
Lichen & 0.851 & 0.848 & 0.841 & 0.832 & \textbf{0.855} & 0.848\\
Mullein & 0.773 & 0.769 & \textbf{0.781} & 0.772 & 0.776 & 0.778\\
Pima & \textbf{0.784} & 0.773 & 0.771 & 0.758 & 0.766 & 0.763\\
\addlinespace
Sonar & 0.726 & 0.736 & 0.774 & \textbf{0.792} & 0.717 & 0.745\\
\em{Time (seconds)} & \em{} & \em{} & \em{} & \em{} & \em{} & \em{}\\
BreastCancer & 6.04 & 2.53 & 2.37 & \textbf{1.45} & 3.69 & 3.69\\
Lichen & 89.7 & 12.8 & 75.1 & \textbf{9.90} & 42.2 & 42.2\\
Mullein & 1,080 & 143 & 1,440 & \textbf{124} & 680 & 680\\
\addlinespace
Pima & 7.94 & 2.91 & 2.55 & \textbf{1.45} & 2.06 & 2.06\\
Sonar & 33.6 & 5.28 & 5.82 & \textbf{2.86} & 11.3 & 11.3\\
\bottomrule
\end{tabular}
\label{tab:enbin}
\end{table}

The following code snippet demonstrates how to tune an elastic net model
for the Boston Housing data using \texttt{glmnet}.

\begin{Shaded}
\begin{Highlighting}[]
\NormalTok{bh\_train}\SpecialCharTok{$}\NormalTok{chas }\OtherTok{\textless{}{-}} \FunctionTok{as.numeric}\NormalTok{(}\FunctionTok{as.character}\NormalTok{(bh\_train}\SpecialCharTok{$}\NormalTok{chas))}
\NormalTok{bh\_test}\SpecialCharTok{$}\NormalTok{chas }\OtherTok{\textless{}{-}} \FunctionTok{as.numeric}\NormalTok{(}\FunctionTok{as.character}\NormalTok{(bh\_test}\SpecialCharTok{$}\NormalTok{chas))}

\NormalTok{foldid }\OtherTok{\textless{}{-}} \FunctionTok{sample}\NormalTok{(}\DecValTok{1}\SpecialCharTok{:}\DecValTok{10}\NormalTok{, }\AttributeTok{size =} \FunctionTok{nrow}\NormalTok{(bh\_train), }\AttributeTok{replace =} \ConstantTok{TRUE}\NormalTok{)}
\NormalTok{alpha }\OtherTok{\textless{}{-}} \FunctionTok{seq}\NormalTok{(}\DecValTok{0}\NormalTok{, }\DecValTok{1}\NormalTok{, }\FloatTok{0.1}\NormalTok{)}
\NormalTok{alpha\_data }\OtherTok{\textless{}{-}} \FunctionTok{data.frame}\NormalTok{(}\AttributeTok{alpha =}\NormalTok{ alpha, }\AttributeTok{lambda =} \ConstantTok{NA}\NormalTok{, }\AttributeTok{loss =} \ConstantTok{NA}\NormalTok{)}
\NormalTok{model\_cv }\OtherTok{\textless{}{-}} \ConstantTok{NULL}
\ControlFlowTok{for}\NormalTok{ (i }\ControlFlowTok{in} \DecValTok{1}\SpecialCharTok{:}\FunctionTok{length}\NormalTok{(alpha)) \{}
\NormalTok{  model\_cv[[i]] }\OtherTok{\textless{}{-}} \FunctionTok{cv.glmnet}\NormalTok{(}\AttributeTok{x =} \FunctionTok{as.matrix}\NormalTok{(}\FunctionTok{subset}\NormalTok{(bh\_train, }\AttributeTok{select =} \SpecialCharTok{{-}}\NormalTok{cmedv)), }
                             \AttributeTok{y =}\NormalTok{ bh\_train}\SpecialCharTok{$}\NormalTok{cmedv, }\AttributeTok{family =} \StringTok{"gaussian"}\NormalTok{, }
                             \AttributeTok{type.measure =} \StringTok{"mse"}\NormalTok{)}
\NormalTok{  alpha\_data[i, }\SpecialCharTok{{-}}\DecValTok{1}\NormalTok{] }\OtherTok{\textless{}{-}} \FunctionTok{c}\NormalTok{(model\_cv[[i]]}\SpecialCharTok{$}\NormalTok{lambda.min, }
\NormalTok{                         model\_cv[[i]]}\SpecialCharTok{$}\NormalTok{cvm[model\_cv[[i]]}\SpecialCharTok{$}\NormalTok{lambda }\SpecialCharTok{==} 
\NormalTok{                                             model\_cv[[i]]}\SpecialCharTok{$}\NormalTok{lambda.min])}
\NormalTok{\}}

\NormalTok{model }\OtherTok{\textless{}{-}} \FunctionTok{glmnet}\NormalTok{(}\AttributeTok{x =} \FunctionTok{as.matrix}\NormalTok{(}\FunctionTok{subset}\NormalTok{(bh\_train, }\AttributeTok{select =} \SpecialCharTok{{-}}\NormalTok{cmedv)), }
                \AttributeTok{y =}\NormalTok{ bh\_train}\SpecialCharTok{$}\NormalTok{cmedv, }\AttributeTok{family =} \StringTok{"gaussian"}\NormalTok{,}
                \AttributeTok{lambda =}\NormalTok{ alpha\_data}\SpecialCharTok{$}\NormalTok{lambda[alpha\_data}\SpecialCharTok{$}\NormalTok{loss }\SpecialCharTok{==} 
                                             \FunctionTok{min}\NormalTok{(alpha\_data}\SpecialCharTok{$}\NormalTok{loss)][}\DecValTok{1}\NormalTok{],}
                \AttributeTok{alpha =}\NormalTok{ alpha\_data}\SpecialCharTok{$}\NormalTok{alpha[alpha\_data}\SpecialCharTok{$}\NormalTok{loss }\SpecialCharTok{==} 
                                           \FunctionTok{min}\NormalTok{(alpha\_data}\SpecialCharTok{$}\NormalTok{loss)][}\DecValTok{1}\NormalTok{],}
                \AttributeTok{type.measure =} \StringTok{"mse"}\NormalTok{)}

\NormalTok{result }\OtherTok{\textless{}{-}} \FunctionTok{predict}\NormalTok{(model, }\FunctionTok{as.matrix}\NormalTok{(}\FunctionTok{subset}\NormalTok{(bh\_test, }\AttributeTok{select =} \SpecialCharTok{{-}}\NormalTok{cmedv)), }
                  \AttributeTok{type =} \StringTok{"response"}\NormalTok{)}
\NormalTok{rmse.en }\OtherTok{\textless{}{-}} \FunctionTok{rmse\_vec}\NormalTok{(}\AttributeTok{truth =}\NormalTok{ bh\_test}\SpecialCharTok{$}\NormalTok{cmedv, }\AttributeTok{estimate =}\NormalTok{ result[, }\DecValTok{1}\NormalTok{])}
\NormalTok{mae.en }\OtherTok{\textless{}{-}} \FunctionTok{mae\_vec}\NormalTok{(}\AttributeTok{truth =}\NormalTok{ bh\_test}\SpecialCharTok{$}\NormalTok{cmedv, }\AttributeTok{estimate =}\NormalTok{ result[, }\DecValTok{1}\NormalTok{])}

\FunctionTok{data.frame}\NormalTok{(}\AttributeTok{RMSE =}\NormalTok{ rmse.en, }\AttributeTok{MAE =}\NormalTok{ mae.en)}
\end{Highlighting}
\end{Shaded}

The following code snippet shows how to tune an elastic net model for
the Boston Housing data using \texttt{EZtune} with the genetic algorithm
and the fast option.

\begin{Shaded}
\begin{Highlighting}[]
\NormalTok{model }\OtherTok{\textless{}{-}} \FunctionTok{eztune}\NormalTok{(}\AttributeTok{x =} \FunctionTok{subset}\NormalTok{(bh\_train, }\AttributeTok{select =} \SpecialCharTok{{-}}\NormalTok{cmedv), }\AttributeTok{y =}\NormalTok{ bh\_train}\SpecialCharTok{$}\NormalTok{cmedv,}
                \AttributeTok{method =} \StringTok{"en"}\NormalTok{, }\AttributeTok{optimizer =} \StringTok{"ga"}\NormalTok{, }\AttributeTok{fast =} \FloatTok{0.5}\NormalTok{)}

\NormalTok{predictions }\OtherTok{\textless{}{-}} \FunctionTok{predict}\NormalTok{(model, bh\_test)}
\NormalTok{rmse.ez }\OtherTok{\textless{}{-}} \FunctionTok{rmse\_vec}\NormalTok{(}\AttributeTok{truth =}\NormalTok{ bh\_test}\SpecialCharTok{$}\NormalTok{cmedv, }\AttributeTok{estimate =}\NormalTok{ predictions)}
\NormalTok{mae.ez }\OtherTok{\textless{}{-}} \FunctionTok{mae\_vec}\NormalTok{(}\AttributeTok{truth =}\NormalTok{ bh\_test}\SpecialCharTok{$}\NormalTok{cmedv, }\AttributeTok{estimate =}\NormalTok{ predictions)}

\FunctionTok{data.frame}\NormalTok{(}\AttributeTok{RMSE =}\NormalTok{ rmse.ez, }\AttributeTok{MAE =}\NormalTok{ mae.ez)}
\end{Highlighting}
\end{Shaded}

Table~\ref{tab:enreg} shows the mean RMSE and computation times for the regression
elastic net models. As with binary classification, there is no one
method that out performs the others. The table also shows the
\texttt{glmnet} method was faster for regression than the other
datasets, but all of them were fast.

\begin{table}

\caption{Mean RMSE and computation times in seconds for ten trials of tuning regression elastic net models. The best results for each dataset are bolded.}
\centering
\begin{tabular}[t]{l>{}l>{}l>{}l>{}l>{}l>{}l}
\toprule
\multicolumn{1}{c}{ } & \multicolumn{4}{c}{EZtune} & \multicolumn{2}{c}{Glmnet} \\
\cmidrule(l{3pt}r{3pt}){2-5} \cmidrule(l{3pt}r{3pt}){6-7}
Data & GA CV & GA fast & HJ CV & HJ fast & 1-SE & Min\\
\midrule
\em{RMSE} & \em{} & \em{} & \em{} & \em{} & \em{} & \em{}\\
Abalone & 2.23 & \textbf{2.19} & 2.21 & 2.22 & 2.30 & 2.22\\
BostonHousing & 4.87 & 4.86 & 4.84 & 4.77 & 4.87 & \textbf{4.63}\\
CO2 & 6.07 & 7.34 & 6.53 & 6.64 & 6.51 & \textbf{6.01}\\
Crime & \textbf{24.4} & 30.2 & 28.3 & 27.6 & 27.9 & 27.6\\
\addlinespace
\em{Time (seconds)} & \em{} & \em{} & \em{} & \em{} & \em{} & \em{}\\
Abalone & 5.71 & 2.67 & 2.20 & \textbf{1.42} & 1.66 & 1.66\\
BostonHousing & 5.14 & 2.35 & 1.97 & 1.12 & \textbf{0.918} & \textbf{0.918}\\
CO2 & 5.15 & 2.45 & 1.85 & 1.27 & \textbf{0.737} & \textbf{0.737}\\
Crime & 5.13 & 2.51 & 1.83 & 1.16 & \textbf{0.777} & \textbf{0.777}\\
\bottomrule
\end{tabular}
\label{tab:enreg}
\end{table}

\hypertarget{model}{%
\section{Model selection with EZtune}\label{model}}

\begin{figure} 
    \centering
    \includegraphics[width=1.0\textwidth]{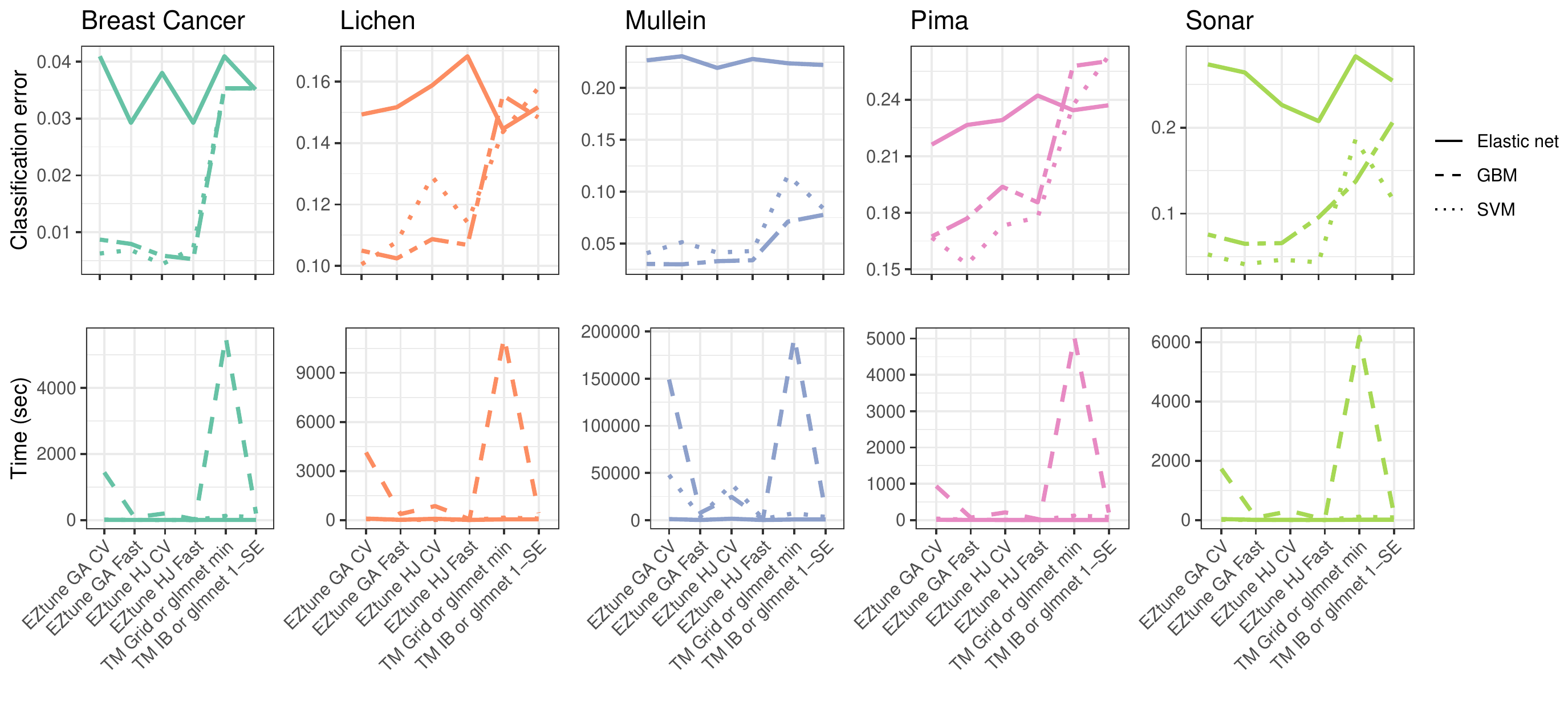}
    \caption{Classification errors and computation times for datasets with a
binary response.}    
    \label{fig:binmod}
\end{figure}

As stated earlier, there is no one model type that out performs other
models in all situations \cite{schumacher2001no}. Thus,
different model types should be compared when developing a model.
\texttt{EZtune} provides an easy interface for comparing different
models. Figure~\ref{fig:binmod} shows the mean classification errors and
mean computation times for ten models tuned with \texttt{EZtune},
\texttt{tidymodels}, and \texttt{glmnet} for all five of the binary
classification datasets. SVM performed better for some datasets and GBM
for others. The best type of model also depends on the method that was
used tune the model. GBM and SVM performed similarly well for the Breast
Cancer data with the SVM performing slightly better for most of the
models. In many cases, the GBM and SVM models are comparable. However,
for some datasets, one of the models consistently outperforms the
others. For example, the model with the lowest classification error for
the Sonar data is an SVM tuned with \texttt{EZtune}, while the best
model for the Mullein data is a GBM tuned with \texttt{EZtune}. Not only
is the model type (GBM or SVM) important, different tuning methods
produce models with very different accuracies as is seen with the
Lichen, Pima, Abalone, and Boston Housing datasets. The elastic net
models have greater classification error than the SVM and GBM in nearly
all cases.

\begin{figure} 
    \centering
    \includegraphics[width=1.0\textwidth]{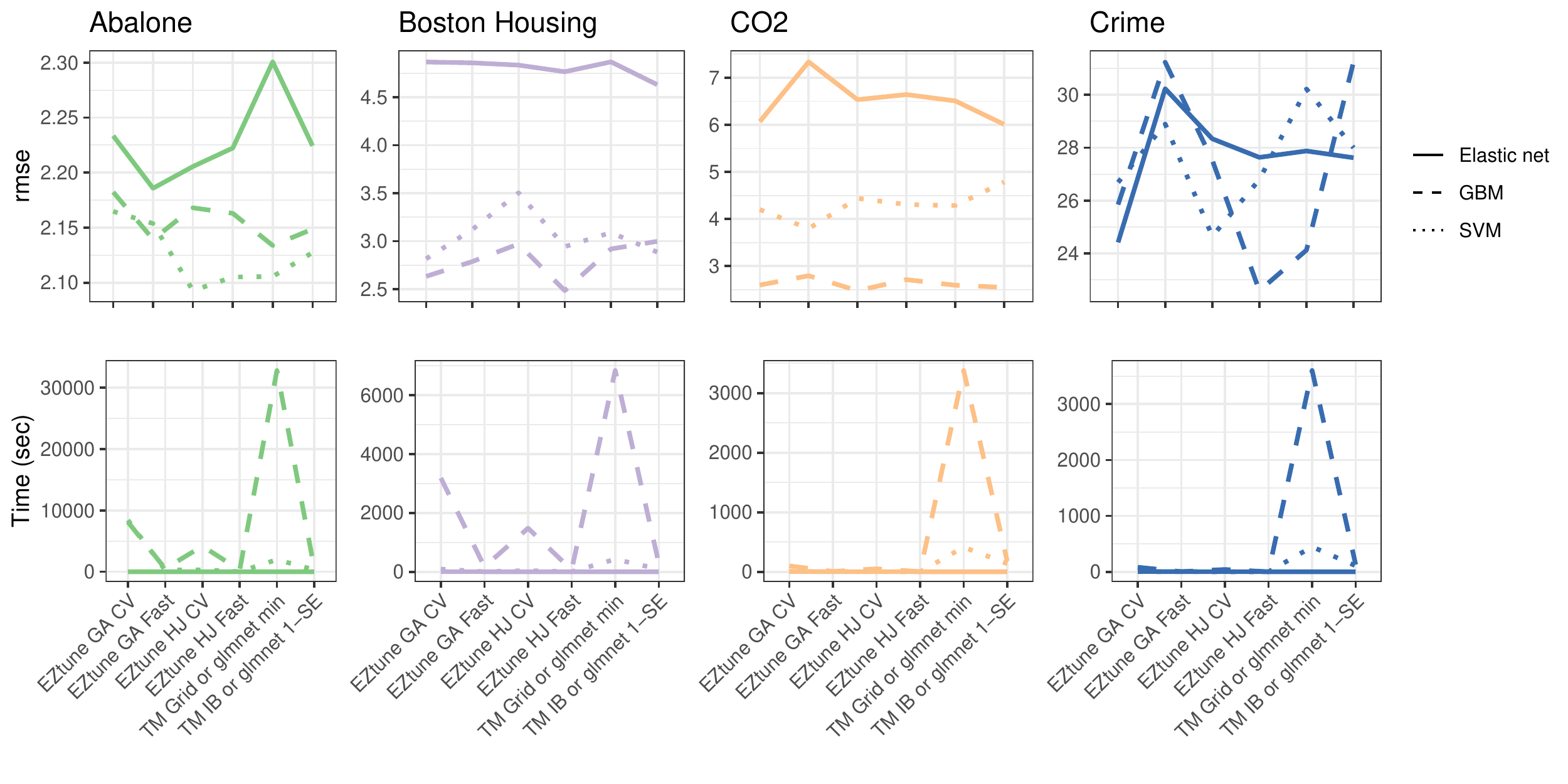}
    \caption{RMSEs and computation times for datasets with a
continuous response.}    
    \label{fig:regmod}
\end{figure}

Figure~\ref{fig:regmod} shows the RMSE and computation time for the
regression models. Figure~\ref{fig:binmod} and Figure~\ref{fig:regmod}
show that the elastic net model had a larger error rate than SVM and GBM
for all of the datasets with the exception of the Crime dataset. The
Crime dataset is very small and hyperparameter tuning is difficult with
small datasets \cite{dissertation}. The SVM models performed better for the
Abalone data, but the GBM was the better model for the Boston Housing
and CO2 datasets.

\hypertarget{conclusions}{%
\section{Conclusions}\label{conclusions}}

Supervised learning models have the ability to increase prediction
accuracy if they are tuned well, but tuning such models is not trivial.
We discussed the advantages and disadvantages using \texttt{caret},
\texttt{tidymodels}, and \texttt{EZtune} for tuning such models. All of
these packages are good options for tuning, but they each have strengths
and weaknesses. Experienced R users may prefer the power and flexibility
of \texttt{caret}, but that flexibility comes with the price of being
difficult to learn and implement, even for experienced R users. Further,
\texttt{tidymodels} is a good option for experienced R users who have a
solid understanding of hyperparameters and wish to explore a larger set
of statistical learning models. In contrast, \texttt{EZtune} is an
excellent option for users who want fast and effective hyperparameter
tuning for a smaller set of model types without the programming overhead
required for other approaches. \texttt{EZtune} is a powerful tuning tool
whose simple interface, ability to find a well tuned model, and fast
computation time make it an excellent choice for general hyperparameter
tuning or incorporation into a larger computational pipeline. Not only
is \texttt{EZtune} an approachable option for someone new to statistical
learning models, it is an excellent way to become familiar with
statistical learning models and their hyperparameters. \texttt{EZtune}
can be used to prepare users to interact with \texttt{tidymodels} and
\texttt{caret} in the future if they choose to expand their choice of
models.

\bibliographystyle{unsrt}  
\bibliography{eztune}  






\end{document}